\documentclass[runningheads]{llncs}

\usepackage{eccv}

\usepackage{eccvabbrv}

\usepackage{graphicx}
\usepackage{booktabs}

\usepackage[accsupp]{axessibility}  

\usepackage[pagebackref,breaklinks,colorlinks,citecolor=eccvblue]{hyperref}

\usepackage{orcidlink}

\def\mmethod{AggDet}
\usepackage{multirow}
\usepackage{enumitem}
\usepackage{mathtools}
\usepackage{wrapfig}
\usepackage{appendix}

\begin{document}

\title{Training-free Boost for Open-Vocabulary Object Detection with Confidence Aggregation} 

\titlerunning{AggDet}

\author{Yanhao Zheng$^{*}$, 
Kai Liu$^{*\dagger}$}

\authorrunning{YH.~Zheng et al.}

\institute{
Zhejiang University 
\\
~\\
$^{*}$ Euqal contribution. \\
$^{\dagger}$ Corresponding: \email{kail@zju.edu.cn}
}

\maketitle

\begin{abstract}
Open-vocabulary object detection (OVOD) aims at localizing and recognizing visual objects from novel classes unseen at the training time.
Whereas, empirical studies reveal that advanced detectors generally assign lower scores to those novel instances, which are inadvertently suppressed during inference by commonly adopted greedy strategies like Non-Maximum Suppression (NMS), leading to sub-optimal detection performance for novel classes. 
This paper systematically investigates this problem with the commonly-adopted two-stage OVOD paradigm.
Specifically, in the region-proposal stage, proposals that contain novel instances showcase lower objectness scores, since they are treated as background proposals during the training phase.
Meanwhile, in the object-classification stage, novel objects share lower region-text similarities (\ie, classification scores) due to the biased visual-language alignment by seen training samples.
To alleviate this problem, this paper introduces two advanced measures to adjust confidence scores and conserve erroneously dismissed objects: (1) a class-agnostic localization quality estimate via overlap degree of region/object proposals, and (2) a text-guided visual similarity estimate with proxy prototypes for novel classes.
Integrated with adjusting techniques specifically designed for the region-proposal and object-classification stages, this paper derives the \textit{agg}regated confidence estimate for the open-vocabulary object \textit{det}ection paradigm (\mmethod).
Our \mmethod~is a generic and training-free post-processing scheme, which consistently bolsters open-vocabulary detectors across model scales and architecture designs. 
For instance, \mmethod~receives 3.3\% and 1.5\% gains on OV-COCO and OV-LVIS benchmarks respectively, without any training cost.
Extensive experiments and ablation studies are conducted to substantiate \mmethod's effectiveness.
Code is available at \url{https://github.com/WarlockWendell/AggDet}.
\end{abstract}
\section{Introduction}
\label{sec:intro}

Open-vocabulary object detection (OVOD) has witnessed remarkable progress in recent years~\cite{bansal2018zero,gu2021open,zareian2021open,feng2022promptdet,ma2024codet,jin2024llms}.
In the literature, two-stage approaches have shown great potential~\cite{zhou2022detic,kaul2023multi,ma2024codet}: they first generate class-agnostic region proposals, and then utilize advanced vision-language models (VLMs) to classify objects for desired categories.
Largely attributed to VLMs' generalized vision-language alignment capability~\cite{radford2021learning,li2022grounded}, open-vocabulary detectors are able to detect novel classes even not seen during training.
However, a significant performance gap between novel and base classes still exists in recent detectors, and we argue that it is caused by the above two inference stages jointly.

\begin{figure}[ht]
    \centering
    \vspace{-6pt}
    \includegraphics[width=0.8\linewidth]{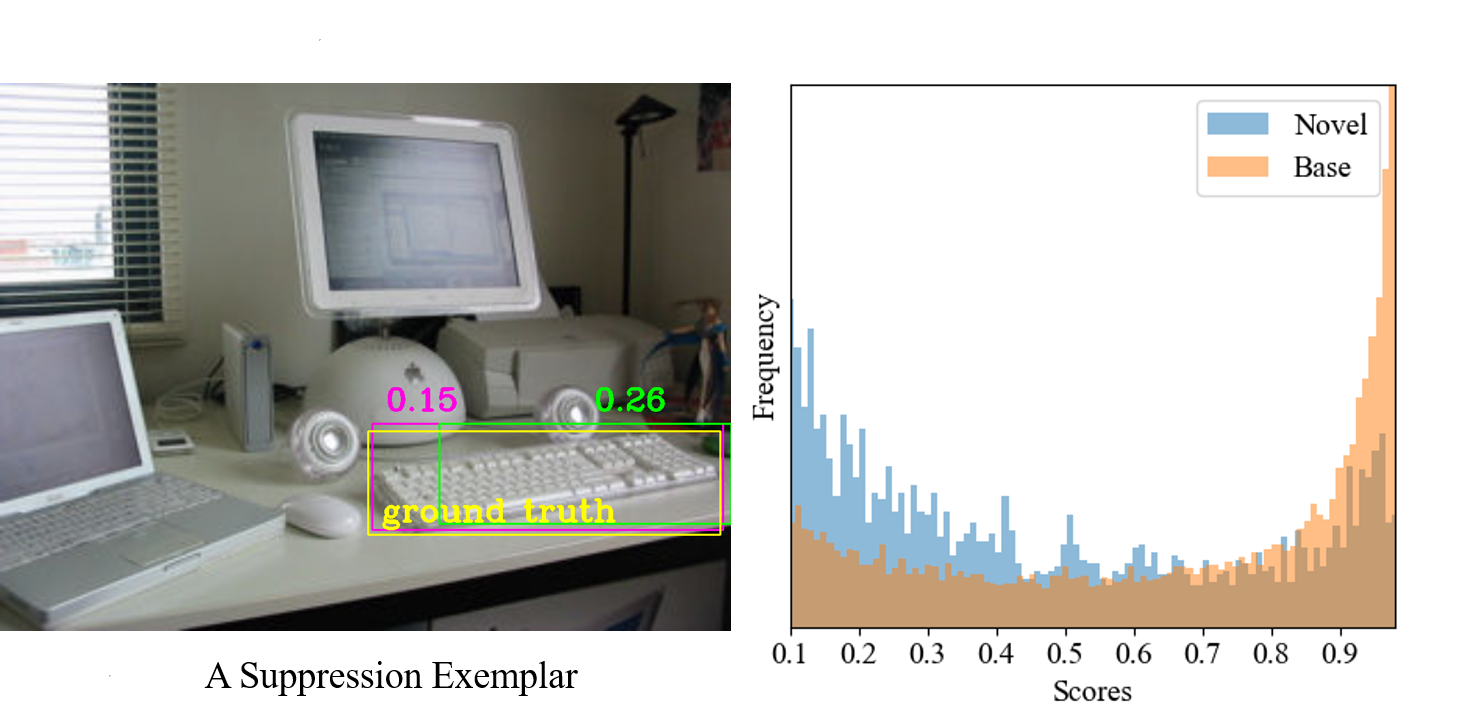}
    \caption{\textbf{Left}: the region-proposal containing novel objects (colored in \textcolor{magenta}{pink}) is suppressed due to its lower objectness score (0.15). \textbf{Right}: detectors generally assign lower classification scores for base classes than novel classes.}
    \label{fig:intro_moti}
\end{figure}

As demonstrated in \cref{fig:intro_moti}, in the first region-proposal stage~\cite{ren2015faster}, proposals that contain novel instances are assigned by lower objectness scores.
The primary reason is that during training, novel proposals are treated as background and the objectness scores are consequently suppressed~\cite{zhou2022detic,ma2024codet}.
As a result, they are inadvertently filtered out by the NMS strategy, and the detection recall for novel classes is declined.
Meanwhile, \cref{fig:intro_moti} reveals that in the second object-classification stage, novel objects obtain much lower confidence scores (\ie, region-text similarities) compared with base instances.
This is mainly because the fine-tuned visual-language feature space is inevitably over-fitted on base (seen) categories during training~\cite{jin2024llms}.
Therefore, novel objects are further dismissed by thresholding strategies (such as Top-K selection~\cite{zhou2022detic}), leading to sub-optimal detection performance for novel classes.

To conserve the erroneously dismissed novel objects, we propose to aggregate the inference-time confidence estimates from two main aspects.

First, as the objectness confidence is unreliable for region proposals, we integrate it with class-agnostic localization quality estimates for NMS.
Specifically, we quantify the degree to which the proposals overlap with each other to measure the localization quality.
It is inspired by~\cite{Tang2018PCLPC}, which indicates that high-quality proposals tend to be clustered and overlapped around the objects they contain, regardless of their categories.
Proposals for novel classes with high localization quality will be preserved, even if the objectness confidence is relatively lower.

Then, since the region-text similarities are biased to base categories, we aggregate the object-classification confidence for novel classes with region-prototype similarities.
In particular, we establish the visual prototypes for novel classes by employing the delta-consistency property of VLMs' feature space\cite{lyu2023deltaedit}, guided by textual prompts from base and novel classes.
If the semantic distance between objects' feature representations and the visual prototypes is close enough, the objects from novel categories will be conserved even if they share relatively lower region-text similarities.
In addition, we leverage the localization quality estimates in this stage as well, so as to obtain a better object confidence aggregation.

This paper ultimately induces a generic and training-free post-processing schema with proposed \textit{agg}regation techniques to boost open-vocabulary object \textit{det}ection, termed as \mmethod.
To evaluate the effectiveness, extensive ablations and experiments are conducted on the popular OV-COCO and OV-LVIS benchmarks.
The results show that each proposed technique contributes to the performance gain in an almost orthogonal way, and the ultimate \mmethod~consistently bolsters various open-vocabulary detectors across model scales and architecture design. 
For instance, \mmethod~receives up to 3.3\% and 1.5\% gains on OV-COCO~\cite{lin2014microsoft} and OV-LVIS~\cite{gupta2019lvis} benchmarks respectively, without any training cost.
Additionally, our method only introduces highly optimized multiplication and addition operations for confidence aggregation, which only brings a less than 1 ms latency during inference, addressing the practicality of our \mmethod.

To the best of our knowledge, we are the first to conduct a systematic investigation on confidence aggregation for novel classes during detection inference.
We hope this paper can bring new insights to the OVOD community, and also translate the aggregation techniques into the training paradigm for further enhancement.
Our contribution can be summarized as follows:
\begin{itemize}[itemsep=2pt,topsep=0pt,parsep=0pt]
    \item We investigate the performance gap between novel and base classes for current OVOD detectors, and reveal the bias on confidence estimate at both the two detection stages.
    \item We propose to measure the class-agnostic localization quality, as well as the text-guided prototype similarity, to aggregate the confidence estimates to recognize novel classes.
    \item We derive \mmethod, a generic and training-free schema to boost OVOD with various advanced detectors, which showcases a consistent enhancement for the detection performance by extensive experiments and ablation studies.
\end{itemize}

\section{Related Works}
\label{sec:rel}

\textbf{Closed-Vocabulary Object Detection. }
In general, closed vocabulary object detection algorithms can be categorized into three main paradigms: one-stage detectors\cite{redmon2016you,liu2016ssd} two-stage detectors\cite{ren2015faster, he2017mask,cai2018cascade}, and transformer-based detectors\cite{carion2020end}. One-stage detectors divide the input image into dense regular grids and predict object class probabilities coupled with bounding box regressions for each spatial grid. In contrast, two-stage detectors first extract class-agnostic region proposals from the input image, and obtain region-wise features via region-of-interest pooling or alignment operators (\ie, RoIPool\cite{girshick2014rich} or RoIAlign\cite{he2017mask}), and subsequently feed these pooled features into classification and bounding box regression head. Recently, CenterNet2\cite{zhou2021probabilistic}employed a class-agnostic one-stage detector as the proposal network to construct a probabilistic two-stage detector framework, leading to an efficient sparse proposal generation scheme. Transformer-based detectors reformulate object detection as a set prediction problem, resulting in a more streamlined end-to-end pipeline\cite{carion2020end}. However, most close vocabulary object detection methods rely on human-annotated data heavily and cannot generalize to categories unseen during the training stage.

\textbf{Open-Vocabulary Object Detection (OVOD). }
Leveraging the powerful zero-shot classification ability of large-scale Vision-Language Models(VLMs) such as CLIP\cite{du2022learning} and ALIGN\cite{jia2021scaling}, OVOD can be extended to categories absent from the training set. ViLD\cite{gu2021open} distills knowledge from CLIP into a two-stage detector, with the objective of aligning CLIP’s image feature space with that of the Faster R-CNN backbone to enable open-vocabulary generalization. HierKD\cite{ma2022open} introduce hierarchical knowledge distillation methods to capture both global and local knowledge from VLMs. BORAN\cite{wu2023aligning} employs a Bag of Regions approach to distill knowledge collectively, thereby fully leveraging the scene-understanding capabilities of VLMs. Detic\cite{zhou2022detic} and Rasheed\cite{bangalath2022bridging} utilize weakly supervised methods to augment training data by leveraging datasets annotated at the image level. 
PromptDet\cite{feng2022promptdet} employs prompt learning techniques to better align the textual space and region feature space. MM-OVOD\cite{kaul2023multi} and DVDet\cite{jin2024llms} use large language models (LLMs) to generate more informative descriptors for each class, resulting in improved detection performance. CoDet\cite{ma2024codet} views the alignment between regions and textual descriptions as a co-occurring object discovery problem, thus overcoming the necessity for pre-aligned vision-language spaces. Our work systematically investigates the confidence aggregation for novel classes during detection inference, thereby reducing the performance gap between novel and base categories in the current mainstream two-stage OVOD methods.

\textbf{Post-hoc Enhancement for OVOD. }
In order to narrow the performance gap between Novel and Base categories during testing, some OVOD methods (such as ViLD\cite{du2022learning}, ProxyDet\cite{jeong2023proxydet}) compute not only the model classification scores but also the scores obtained by classifying regions using VLM models. These methods then geometrically average the two scores with different coefficients for Novel and Base categories, leading to improved classification results. Additionally, MEDet\cite{chen2022open} analyzes that under the OVD training strategy, the classification scores for Novel categories are lower than those for Base categories. To address this, MEDet proposes an offline class-wise adjustment method to correct the final scores. Unlike the aforementioned approaches, our method does not require the use of additional VLMs for classification during inference. Furthermore, we analyze both stages of the detector and develop a training-free post-processing strategy for performance enhancement.

\section{Method}
\label{sec:method}

This paper introduces a generic and training-free schema (namely \mmethod) to boost commonly-adopted two-stage detectors for OVOD. 
The overview confidence aggregation paradigm is presented in \cref{sec:method_overview}, and the detailed approaches to enhance the first region-proposal stage and the second object-classification stage are described in \cref{sec:method_rpn} and \cref{sec:method_obj}, respectively.

\subsection{Overview}
\label{sec:method_overview}

\begin{figure}[tb]
    \centering
    \includegraphics[width=0.9\linewidth]{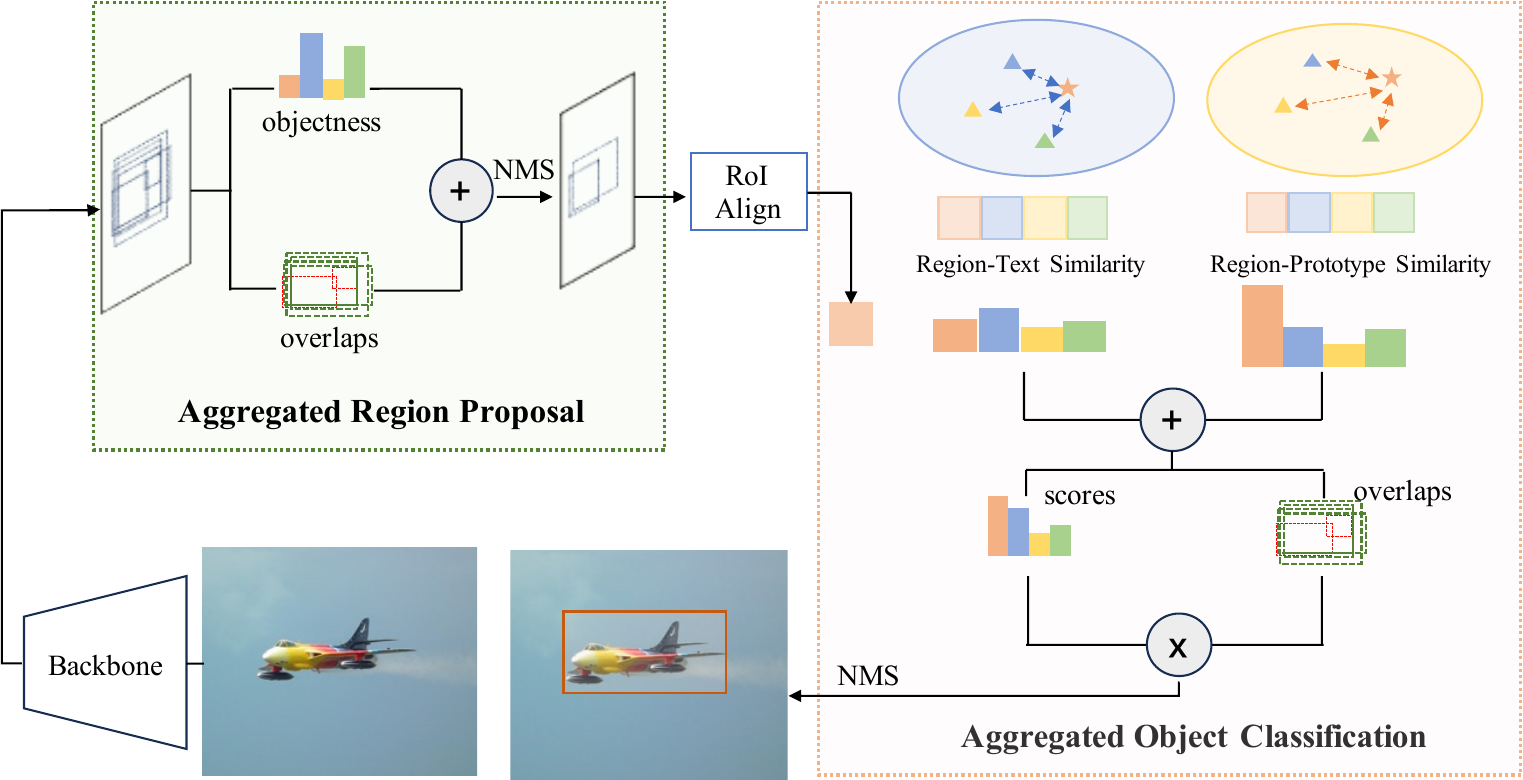}
    \caption{The framework of our proposed \mmethod, which aggregates the confidence estimates at both region-proposal and object-classification stages.}
    \label{fig:framework}
\end{figure}

\cref{fig:framework} illustrates the overview paradigm for out \mmethod. 
Typically, a two-stage detector~\cite{ren2015faster,he2017mask} consists of a feature extraction backbone $\mathcal{F}$, a class-agnostic region-proposal network $\mathcal{T}$, and an open vocabulary object classifier $\Psi$ (by measuring the similarity between object representation and textual embedding for targeted class). 
The detection procedure can be formulated as:

\begin{equation}
    \label{eq:frcnn_fmwk}
    \{\mathbf{y}_i\}_{i=1}^{N} = \Psi \left(\text{NMS} \left(\mathcal{T} \left(\mathcal{F} (\mathbf{x}) \right)\right)\right)
\end{equation}

\noindent where $\{\mathbf{y}_i\}_{i=1}^{N}$ is the output prediction set with the image $\mathbf{x}$ as input, and $\text{NMS}$ refers to the Non-Maximum Suppression\footnote{RoIAlign and R-CNN are hidden for simplicity.}.

In the region-proposal stage, we amplify the proposals' objectness confidence by adding the localization quality estimates with the overlap information. The proposals that contain novel objects and share high quality will be preserved after NMS, even if they are assigned with a lower objectness.
We denote the aggregated region-proposal process as $\mathcal{T}^{Agg}$.
At the object-classification stage, we estimate the proxy visual prototypes for novel categories, and aggregate the vanilla region-text similarities by supplementing the object-prototype similarities.
Meanwhile, the localization quality is employed to further scale the classification confidence.
The aggregated object-classification process is summarized as $\Psi^{Agg}$.
Our ultimate \mmethod~paradigm can be expressed as:

\begin{equation}
    \label{eq:ours_fmwk}
    \{\mathbf{y}^*_i\}_{i=1}^{N} = \Psi^{Agg} \left(\text{NMS} \left(\mathcal{T}^{Agg} \left(\mathcal{F} (\mathbf{x}) \right)\right)\right)
\end{equation}

We now introduce and formulate the detailed implementation of $\mathcal{T}^{Agg}$ and $\Psi^{Agg}$ in the following sections.

\subsection{Aggregated Region Proposal}
\label{sec:method_rpn}

Given the image $\mathbf{x}$ as input, the $M$ region-proposals is generated by $\{(\textbf{b}_i, o_i)\}_{i=1}^{M} = \mathcal{T} (\mathcal{F} (\mathbf{x}))$, where $\textbf{b}_i \in \mathbb{R}^4$ and $o_i \in \mathbb{R}^1$ refer to the bounding-box and objectness for the $i$-th proposal.
After the Non-Maximum Suppression (NMS), the reserved $M^\prime$ proposals are sent to classifier $\Psi$ to produce detection outputs.
At the conventional region-proposal stage, proposals that contain novel objects will receive lower objectness confidence $o_i$ due to the absence during the training phase.
As a result, they are inadvertently suppressed by the NMS process. 

\begin{figure}
  \centering
  \begin{subfigure}{0.48\linewidth}
    \centering
    \includegraphics[width=.9\linewidth]{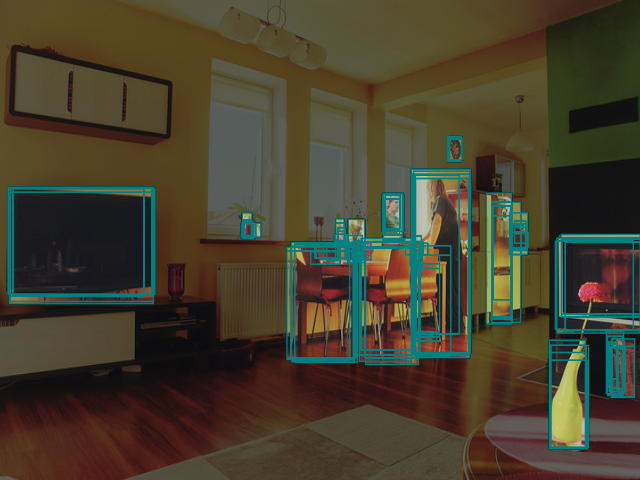}
    \label{fig:short-a}
  \end{subfigure}
  \hfill
  \begin{subfigure}{0.48\linewidth}
    \centering
    \includegraphics[width=.9\linewidth]{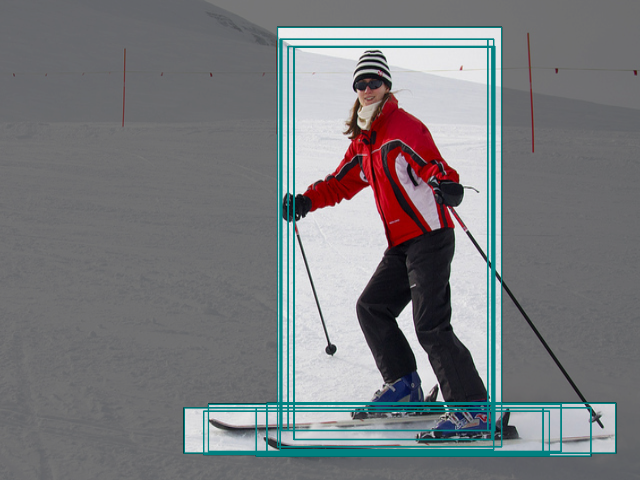}
    \label{fig:example}
  \end{subfigure}
  \caption{An illustration depicting proposals clustered around the ground truth objects.}
  \label{fig:agg_rpn_moti}
\end{figure}

To alleviate this problem, we introduce the aggregated objectness confidence $o^{Agg}_i$ by integrating the localization quality of $\textbf{b}_i$ for the $i$-th region proposal.
This is motivated by \cite{Tang2018PCLPC}, which indicates foreground objects are usually contained by several overlapped region proposals, regardless of the categories that the objects belong to.
Our empirical observation in \cref{fig:agg_rpn_moti} further confirms this hypothesis, meaning that the overlap degree of those region proposals can effectively measure their class-agnostic localization quality.
Thus, we adopt the quality measurement to save the proposals with under-estimated objectness confidence for novel objects.

Specifically, we first calculate the Intersection-over-Union (IoU) matrix $\textbf{P} \in \mathbb{R}^{M \times M}$ to represent the overlap degree among the initial $M$ region-proposals: $\textbf{P}_{i,j} = \text{IoU}(\textbf{b}_i, \textbf{b}_j), \; i,j \in \{1,2,...M\}$.
Then, we pick the top-K proposals sharing the largest overlap degree (IoU) with the $i$-th proposal, and average the IoU values to measure the localization quality $q_i$:

\begin{equation}
    \label{eq:agg_rpn_overlap}
    q_i = \frac{1}{K}\sum_{j}\textbf{P}_{i,j}, \quad \text{where}\; j \in TopK(\textbf{P}_{i,:}) \; \text{and}\; j \neq i
\end{equation}

Furthermore, the original objectness $o_i$ is averaged by $q_i$ to derive the aggregated $o^{Agg}_i$ and $\mathcal{T}^{Agg}$:

\begin{equation}
    \label{eq:agg_rpn_obj}
    o^{Agg}_i = \frac{1}{2}\left(o_i + q_i\right), \quad \mathcal{T}^{Agg} \coloneqq \mathcal{F} (\mathbf{x}) \mapsto \{(\textbf{b}_i, o^{Agg}_i)\}_{i=1}^{M}
\end{equation}

\begin{figure}
  \centering
  \includegraphics[width=0.95\linewidth]{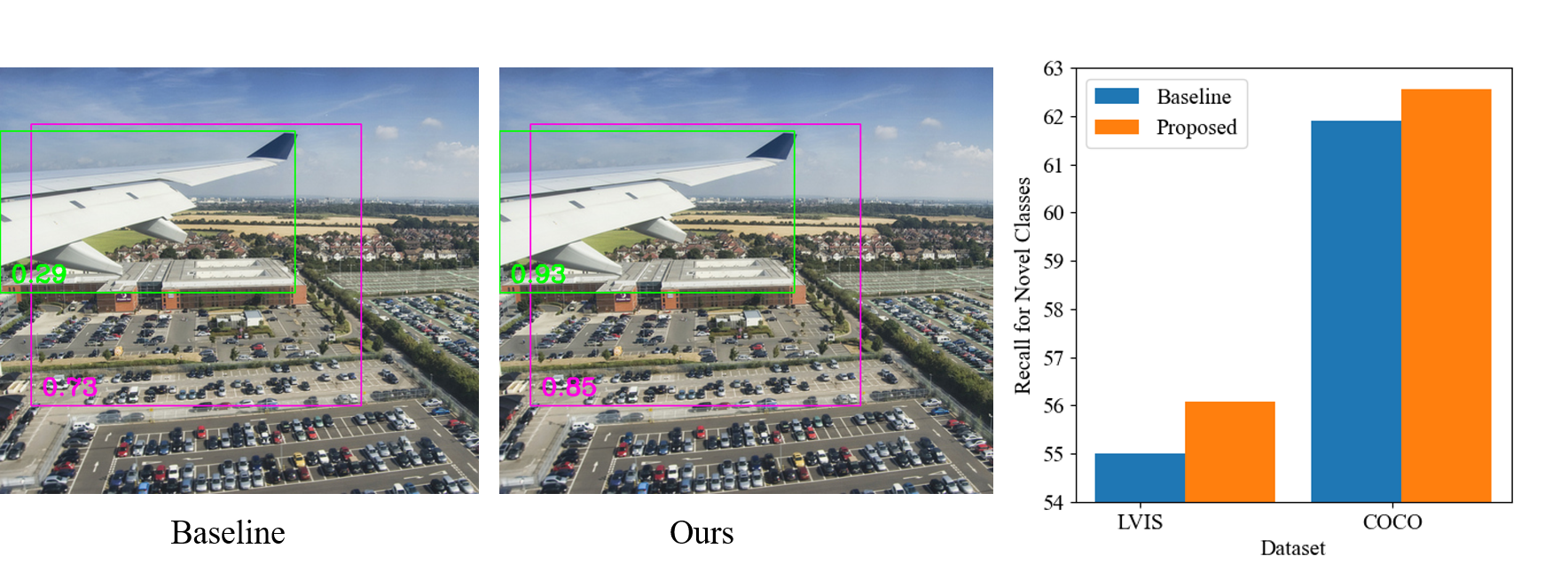}
  \caption{\textbf{Left}: Our aggregated confidence score successfully preserved the validated proposals (colored in \textcolor{green}{green}). \textbf{Right}: the recall difference for novel classes between the baseline method and our proposed \mmethod.  }
  \label{fig:agg_rpn_case}
\end{figure}

By doing so, the region proposals with high localization quality will be preserved after the NMS process, even though the initial objectness confidence is suppressed because of the association with novel objects.
\cref{fig:agg_rpn_case} presents an exemplar to demonstrate its effectiveness.

\subsection{Aggregated Object Classification}
\label{sec:method_obj}
After filtering out $M^\prime$ valid proposals by NMS, the open-vocabulary classifier $\Psi$ will determine their categories by the region-text similarity measurements.
This process can be formulated as $(\tilde{\textbf{b}}_i, \{c_{i,k}\}) = \Psi (\textbf{b}_i, \textbf{f}_i)$, where $\textbf{f}_i$ is the feature representation for the $i$-th proposal, $c_{i,k}$ is the classification confidence that the $i$-th proposal belongs to the $k$-th category, and $\tilde{\textbf{b}}_i$ indicates the refined bounding-box prediction.
In particular, $c_{i,k}$ is estimated by the semantic similarity between visual representation $\textbf{f}_i$ and the text embedding $\textbf{t}_k$ for the $k$-th class, \ie, $c_{i,k} = \sigma(\textbf{f}_i \cdot \textbf{t}_k)$, where ``$\cdot$'' refers to the dot product, $\sigma$ is the \textit{sigmoid} activation, and the temperature $\tau$ is hidden for simplicity.
However, as illustrated before, the region-text similarity for novel classes is underestimated by the biased training samples from base classes, and this paper proposes to mitigate this issue from two main aspects.

\textbf{Visual Similarity Aggregation. }
Even though the generalization capability for visual-language alignment on novel categories declined after fine-tuning, the visual space itself may remain the robust representation abilities.
That is, the visual representation of objects from the same category still exhibits good clustering properties, regardless of the seen and unseen categories~\cite{ma2024codet}.
Therefore, one may find a proxy prototype (clustering centroid) $\textbf{p}_k$ in the visual space for each novel class, and aggregate the underestimated region-text similarity with the well-maintained region-prototype similarity:

\begin{equation}
    \label{eq:agg_cls_sim}
    s^{Agg}_{i,k} = \textbf{f}_i \cdot \textbf{t}_k + \alpha * \textbf{f}_i \cdot \textbf{p}_k, \quad c^{Agg}_{i,k} = \sigma(s^{Agg}_{i,k})
\end{equation}

\noindent where $\alpha$ is the scaling factor.
Note that we only aggregate the region-text similarity for novel classes, but do not alter the well-estimated similarities for base classes.

\begin{wrapfigure}{r}{.5\linewidth}
  \vspace{-16pt}
  \centering
  \includegraphics[width=1.\linewidth]{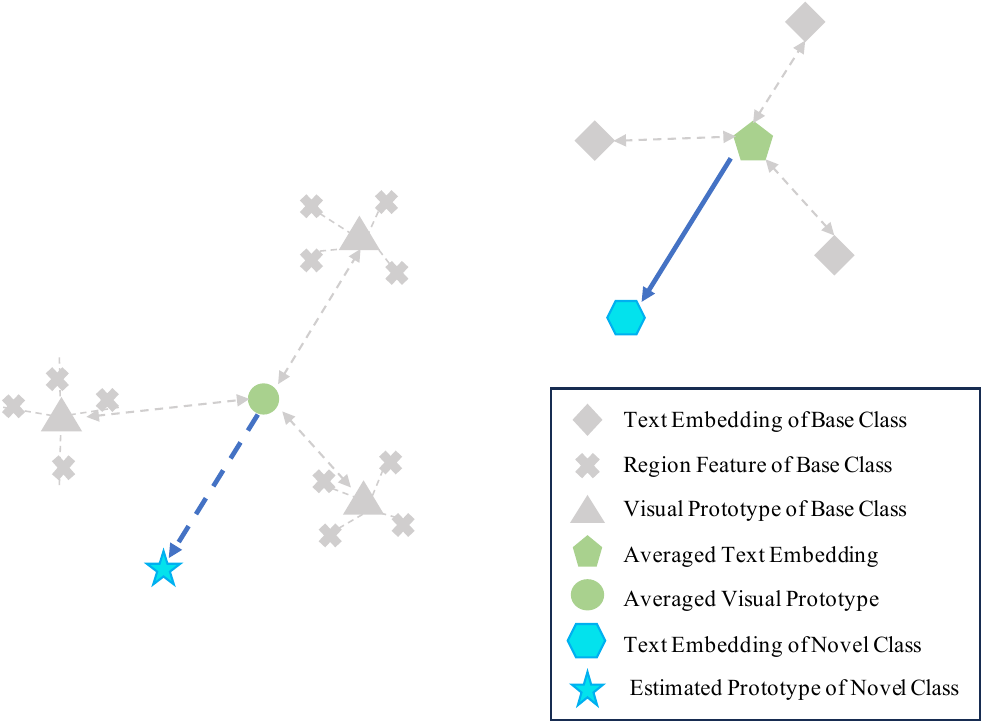}
  \caption{Extrapolation for visual prototypes on novel categories. }
  \label{fig:agg_prototype}
  \vspace{-8pt}
\end{wrapfigure}

As the visual prototype $\textbf{p}_k$ is unavailable for novel categories, we adopt the DeltaEdit~\cite{lyu2023deltaedit} to extrapolate proxy $\hat{\textbf{p}}_k$ using the prototypes from base classes.
This is motivated by the delta-consistency between visual space and language space for VLMs. 
As shown in \cref{fig:agg_prototype}, the delta on visual representations is consistent with the delta on textual embeddings.
Hence, with the textual guidance, we can easily extrapolate the novel prototype from the base prototypes:

\begin{equation}
    \label{eq:agg_prototype_calc}
    \hat{\textbf{p}}_k - \bar{\textbf{p}}
    \approx
    \textbf{t}_k - \bar{\textbf{t}}
\end{equation}

\noindent where $\bar{\textbf{p}} = \frac{1}{C_{\text{base}}}\sum_l{\textbf{p}_l}$ and $\bar{\textbf{t}} = \frac{1}{C_{\text{base}}}\sum_l{\textbf{t}_l}$ are the averaged visual prototypes and text embeddings on all base categories, respectively.
For each base category, we randomly select 300 samples from the training set to calculate the visual prototype $\textbf{p}_l = \frac{1}{300}\sum_j{\textbf{f}_j}$, as it is robust enough to derive the proxy prototypes for novel categories.
We leave the discussion in \cref{sec:app_A}.

Using \cref{eq:agg_prototype_calc}, one may replace the ideal prototype $\textbf{p}_k$ with the approximate prototype $\hat{\textbf{p}}_k$ in \cref{eq:agg_cls_sim}, to justify the classification confidence for the $k$-th novel category.

\textbf{Localization Quality Aggregation. }
Besides the visual similarities to novel prototypes, we found that the localization quality estimates introduced in \cref{sec:method_rpn} also play a vital role in the confidence aggregation at the object-classification stage.
Specifically, the aggregated classification probability $c^{Agg}_{i,*}$ in \cref{eq:agg_cls_sim} is further regulated by:

\begin{equation}
    \label{eq:agg_cls_loc}
    \tilde{c}_{i,*}^{Agg} = \left(c^{Agg}_{i,*}\right)^{\gamma} * \left(q_i\right)^{1 - \gamma}
\end{equation}

Here the localization quality $q_i$ is adopted in case the classification confidence $c^{Agg}_{i,*}$ still shows overestimates on base classes or underestimates on novel classes, and $\gamma$ is a scaling factor.
As $q_i$ is a class-agnostic measurement, we apply \cref{eq:agg_cls_loc} on both novel and base categories. Besides, the calculation differs slightly on the model architecture. For dense proposal generators like Mask-RCNN, we recalculate $q_i$ using the refined boxes predictions $\{\tilde{\textbf{b}}_i\}_{i=1}^{N}$. For sparse proposal models like CenterNet2, we directly employ $q_i$ as computed in \cref{sec:method_rpn}.

By integrating the region-proposal aggregation and object-classification aggregation, this paper ultimately derives the generic and training-free scheme \mmethod~to enhance the open-vocabulary object detection.
Extensive experiments and ablation studies are conducted to evaluate \mmethod's efficacy in the following sections.

\section{Experiments}
\label{sec:exp}

\subsection{Setup}
\label{sec:exp_setup}

\textbf{Datasets. }
Following the OVOD literature~\cite{zhou2022detic,ma2024codet,jin2024llms}, we conduct experiments to evaluate \mmethod’s effectiveness on two prevalent benchmarks, \ie, MS-COCO~\cite{lin2014microsoft} and LVIS-v1.0~\cite{gupta2019lvis}. 
MS-COCO contains 80 categories, of which 48 categories are designated as base (seen) classes, and another 17 classes are adopted as novel (unseen) categories for zero-shot generalization. 
LVIS-v1.0 dataset constitutes a more challenging long-tail distribution with 1203 classes. We treat 337 rare categories as novel classes, with the frequent and common categories as base classes (866 categories in total). 
We denote the open-vocabulary settings for MS-COCO and LVIS as OV-COCO and OV-LVIS, respectively.

\textbf{Metrics. }
We use the mean Average Precision (mAP) at IoU of 0.5 on OV-COCO as the evaluation metric, and report the mask mAP for OV-LVIS.
To thoroughly evaluate our method, results on novel, base (base-c and base-f for OV-LVIS), and all categories are presented respectively.

\textbf{Models. }
This paper mainly investigates the commonly utilized two-stage open-vocabulary object detectors.
To verify the generalization capability and efficacy of the proposed method, we integrate our \mmethod~with various OVOD detectors across model scales and architecture design.
On OV-COCO, we deploy the popular Detic~\cite{zhou2022detic} and recent CoDet~\cite{ma2024codet} with ResNet-50~\cite{he2016deep} (abbreviated as RN50) as the backbone, and the powerful VM-PLM~\cite{zhao2022exploiting} with RN50-FPN as the backbone.
On OV-LVIS, we extend the experiment on Detic and CoDet with Swin-B~\cite{liu2021swin} as the backbone, and adapt to extra detectors like Rasheed~\cite{bangalath2022bridging} and MM-OVOD~\cite{kaul2023multi} with different backbone settings.

\textbf{Implementation Details. }
In most two-stage frameworks (\eg, Mask R-CNN~\cite{he2017mask} and CenterNet2~\cite{zhou2021probabilistic}), our training-free post-processing schema can be directly integrated in the same manner, as described in \cref{sec:method}.
\cref{tab:hyper} displays the hyper-parameters adopted for OV-COCO and OV-LVIS benchmarks.
Despite the slightly different hyper-parameters on the two benchmarks due to their unique data patterns, our \mmethod~still demonstrates the generalization ability across OVOD detectors, as the parameter settings on each benchmark are consistent on different detectors.
The main results are presented and discussed in \cref{sec:exp}, and extensive ablations studies are performed in \cref{sec:abl} to illustrate \mmethod's advances.
Some typical visualizations are provided in \cref{sec:app_B}.

\begin{table}
  \caption{Hyper-parameters for OV-COCO and OV-LVIS benchmarks.}
  \label{tab:hyper}
  \centering
  \begin{tabular}{@{}lccc@{}}
    \toprule
    Dataset & $k$ & $\alpha$ & $\beta$ \\
    \midrule
    OV-COCO & 3 & 0.05 & 3/4 \\
    OV-LVIS & 3 & 0.01 & 2/3 \\
    \bottomrule
  \end{tabular}
\end{table}

\subsection{Main Results}
\label{sec:exp_main}

\textbf{LVIS. }
\cref{tab:res_lvis} shows the performance comparison on the OV-LVIS dataset. 
We integrate our \mmethod~with a series of representative OVOD detectors including Rasheed, CoDet, MM-OVOD, and Detic. 
Both Rasheed and MM-OVOD adopt the variants of ResNet-50 as their backbone, while CoDet and Detic are equipped with the powerful Swin-B for better performance.
The training strategies also differ among those detectors, where CoDet incorporates auxiliary training data from the COCO-Captions~\cite{chen2015microsoft} dataset, and MM-OVOD and Detic leverage the data diversity from ImageNet21k~\cite{deng2009imagenet} to enhance the object representation.
However, despite their different architectures and training configurations, our approach consistently translates to significant performance gains, especially on the mask mAP for novel classes. 
Meanwhile, the detection performance on base classes also slightly increases, which means our class-agnostic localization quality aggregation can benefit both novel and base classes.
The results highlight the efficacy and generalization capability of our \mmethod.
In particular, \mmethod~brings a 1.5\% improvement for the state-of-the-art Detic without any training cost, demonstrating \mmethod's superiority.

\begin{table}[tb]
\caption{Comparison of object detection performance on the OV-LVIS benchmark.}
    \label{tab:res_lvis}
    \centering
    \begin{tabular}{@{}lccccc@{}}
        \toprule
        Method & Backbone & $\text{mAP}_{\text{mask}}^{\text{novel}}$ &$\text{mAP}_{\text{mask}}^{\text{base-c}}$ & $\text{mAP}_{\text{mask}}^{\text{base-f}}$& $\text{mAP}_{\text{mask}}^{\text{all}}$ \\
        \midrule
        ViLD~\cite{gu2021open}       & RN50-FPN & 16.6 & 24.6 & 30.3 & 25.5 \\
        RegionCLIP~\cite{zhong2022regionclip} & RN50     & 17.1 & 27.4 & 36.0 & 28.2 \\
        BARON~\cite{wu2023aligning}      & RN50     & 19.2 & 26.8 & 29.4 & 26.5 \\
        DetPro~\cite{du2022learning}     & RN50     & 19.8 & 25.6 & 28.9 & 25.9 \\
        VLDet~\cite{lin2022vldet}      & RN50     & 21.7 & 29.8 & 34.3 & 30.1 \\
        \midrule
        Rasheed~\cite{bangalath2022bridging} & RN50x8-FPN & 21.1 & 25.0 & 29.1 & 25.9 \\
        Rasheed + AggDet & RN50x8-FPN & \textbf{22.5} & \textbf{25.1} & \textbf{29.1} & \textbf{26.2} \\
        \midrule
        MM-OVOD~\cite{kaul2023multi} & RN50x4 & 27.2 & 33.2 & 35.6 & 33.1\\
        MM-OVOD + AggDet & RN50x4 & \textbf{27.8} & \textbf{33.5} & \textbf{35.6} & \textbf{33.3} \\
        \midrule
        CoDet~\cite{ma2024codet} & Swin-B & 29.4 & 39.5 & 43.0 & 39.2 \\
        CoDet + AggDet & Swin-B & \textbf{30.3} & \textbf{39.6} &  \textbf{43.0} & \textbf{39.4} \\
        \midrule
        Detic~\cite{zhou2022detic} & Swin-B & 33.8 & 41.3 & 42.9 & 40.7\\
        Detic+ AggDet & Swin-B & \textbf{35.3} & \textbf{41.4} & \textbf{43.0} & \textbf{40.9} \\
        \bottomrule
    \end{tabular}
\end{table}

\textbf{OV-COCO. }
\cref{tab:res_coco} further illustrates the consistent performance enhancement by our proposed approach on recent OVOD detectors.
On the OV-COCO benchmark, we investigate three prevalent open vocabulary detection models: CoDet, Detic, and VL-PLM. 
CoDet and Detic are based on the Faster R-CNN framework with a ResNet50-C4 as the backbone, while VL-PLM leverages the ResNet50-FPN architecture to generate multi-scale features. 
According to \cref{tab:res_coco}, our training-free post-processing schema \mmethod~consistently improves the detection of novel classes across all methods (up to 3.3\% gains with Detic). 
In particular, as discussed in \cite{ma2024codet}, CoDet is inferior to the advanced VLDet~\cite{lin2022vldet} model because of the data bias in COCO-Captions, while our \mmethod~directly enables CoDet to overcome this bias and outperform VLDet by a large margin (\ie, 33.2\% \vs 32.0\%), without any training cost.
In addition, \mmethod~further promotes the state-of-the-art detector VL-PLM~\cite{zhao2022exploiting} to higher performance, illustrating the efficacy and potential to boost larger and better OVOD detectors.

\begin{table}[tb]
\caption{Comparison of object detection performance on the OV-COCO benchmark.}
    \label{tab:res_coco}
    \centering
    \begin{tabular}{@{}lcccc@{}}
        \toprule
        Method & Backbone & $\text{mAP}_{\text{50}}^{\text{novel}}$ & $\text{mAP}_{\text{50}}^{\text{base}}$ & $\text{mAP}_{\text{50}}^{\text{all}}$ \\
        \midrule
        RegionCLIP~\cite{zhong2022regionclip} &  RN50  & 26.8 & 54.8 & 47.5 \\
        ViLD~\cite{gu2021open}       &  RN50  & 27.6 & 59.5 & 51.3 \\
        VLDet~\cite{lin2022vldet}      &  RN50  & 32.0 & 50.6 & 45.8 \\
        BARON~\cite{wu2023aligning}      &  RN50  & 33.1 & 54.8 & 49.1 \\
        \midrule
        Detic~\cite{zhou2022detic} & RN50  & 27.8 &  51.1 & 45.0 \\
        Detic + AggDet & RN50 & \textbf{31.1} & \textbf{51.1} & \textbf{45.9}  \\
        \midrule
        CoDet~\cite{ma2024codet} & RN50 & 30.6 & 52.3 & 46.6 \\
        CoDet + AggDet & RN50 & \textbf{33.2} & \textbf{52.4} & \textbf{47.4} \\
        \midrule
        VL-PLM~\cite{zhao2022exploiting} & RN50-FPN & 34.4 & 60.2 & 53.5\\
        VL-PLM + AggDet & RN50-FPN & \textbf{35.0} & \textbf{60.3} & \textbf{53.7} \\
        \bottomrule
    \end{tabular}
\end{table}

\subsection{Ablation Studies}
\label{sec:abl}

In this section, we conduct extensive ablation studies on the OV-COCO and OV-LVIS datasets, using Detic as the baseline detector. 

\textbf{Effectiveness of our proposed techniques. }
To isolate the impact of each proposed modules, we perform an exhaustive permutation and combination of those modules with the Detic baseline, where the hyper-parameters are adopted from \cref{tab:hyper}.
The results are presented in \cref{tab:abl_res}. 
Specifically, we first equip Detic with the localization quality (denoted as LQ) estimates at the aggregated region-proposal (denoted as ARP) stage, and the mAP on novel classes immediately arise on both OV-COCO (+0.4\%) and OV-LVIS (+0.7\%) benchmarks.
Then, we separately integrate Detic with our visual similarity (denoted as VS) and localization quality estimates at the aggregated object-classification (denoted as AOC) stage, and performance improvements were achieved on both datasets to varying degrees.
After combining the proposed techniques, the Detic receives a nearly orthogonal enhancement, especially on novel classes.
For instance, the LQ in the ARP stage and the VS in the AOC stage individually bring 0.4\% and 0.8\% gains on the OV-COCO dataset, and their combination leads to a surprisingly exact boost of 1.2\%.
After incorporating all the proposed modules, our \mmethod~reaches the best performance on both benchmarks.
We conduct further ablations to delve into the in-depth functionality of \mmethod, as discussed in the following sections.

\begin{table}[ht]
\caption{Ablations studies on the efficacy of our proposed techniques.}
    \label{tab:abl_res}
    \centering
    \begin{tabular}{@{}ccc|cc|cccc@{}}
        \toprule
        ARP & \multicolumn{2}{c}{AOC} & \multicolumn{2}{|c|}{OV-COCO} & \multicolumn{4}{c}{OV-LVIS} \\
        \cmidrule(r){1-1} \cmidrule(lr){2-3}  \cmidrule(lr){4-5} \cmidrule(l){6-9}
        LQ & VS & LQ & $\text{mAP}_{\text{50}}^{\text{novel}}$ & $\text{mAP}_{\text{50}}^{\text{all}}$ &  $\text{mAP}_{\text{mask}}^{\text{novel}}$ &$\text{mAP}_{\text{mask}}^{\text{base-c}}$ & $\text{mAP}_{\text{mask}}^{\text{base-f}}$& $\text{mAP}_{\text{mask}}^{\text{all}}$  \\
        \midrule
        & & & 27.8 & 45.0 & 33.8 & 41.3 & 42.9 & 40.7 \\
        \checkmark & & & 28.2 & 45.1 & 34.5 & 41.2 & 42.9 & 40.7 \\
        & \checkmark & & 28.6 & 45.1 & 34.1 & 41.3 & 42.9 & 40.7 \\
        & & \checkmark & 29.3 & 45.4 & 34.0 & 41.3 & 42.9 & 40.6 \\
        \checkmark & \checkmark & & 29.0 & 45.3 & 34.6 & 41.3 & 43.0 & 40.8 \\
        \checkmark & & \checkmark & 29.5 & 45.6 & 34.6 & 41.3 & 42.8 & 40.7 \\
        & \checkmark & \checkmark & 30.9 & 45.8 & 34.8 & 41.2 & 42.8 & 40.8 \\
        \checkmark & \checkmark & \checkmark & \textbf{31.1} & \textbf{45.9} & \textbf{35.3} & \textbf{41.4} & \textbf{43.0} & \textbf{40.9} \\
        \bottomrule
    \end{tabular}
\end{table}

\textbf{Further ablations on region-proposal aggregation. }
In the region-proposal stage, we propose to aggregate the objectness confidence before NMS, as described in \cref{sec:method_rpn}.
In particular, we conduct an ablation study to investigate the efficacy of aggregated confidence $o^{Agg}$ (see \cref{eq:agg_rpn_obj}) by simultaneously reporting the maximal recall $\hat{\text{R}}^{\text{novel}}$ and $\hat{\text{R}}^{\text{all}}$ for novel and all categories.
The maximal recall $\hat{\text{R}}$ measures whether a ground truth object shares an over 50\% IoU with at least a valid region-proposal (the confidence score should exceed 0.1).
According to \cref{tab:res_ablations_rpn}, after replacing the baseline objectness $o$ with the estimated localization quality $q$ for NMS, the maximal recall slightly increases on novel classes while significantly decreases on base classes, leading to a degradation on the maximal recall on all classes.
The result indicates that objectness scores still play a vital role in region-proposal selection, which cannot be easily replaced with localization quality.
Thus, we derive our aggregated confidence $o^{Agg}$, which improves $\hat{\text{R}}^{\text{novel}}$ while maintains $\hat{\text{R}}^{\text{all}}$.
However, the detection performance does not improve with $o^{Agg}$, indicating that further aggregation is required to adjust the detection process in the following stage.

\begin{table}[ht]
\caption{Ablations on localization quality aggregation for the region-proposal stage.}
    \label{tab:res_ablations_rpn}
    \centering
    \begin{tabular}{@{}c|cc|cc@{}}
        \toprule
        Confidence Estimate & $\text{mAP}_{\text{50}}^{\text{novel}}$ & $\text{mAP}_{\text{50}}^{\text{all}}$ & $\hat{\text{R}}^{\text{novel}}$ & $\hat{\text{R}}^{\text{all}}$ \\
        \midrule
        Objectness ($o$)       & 27.8 & 45.0 & 62.0 & 70.0 \\
        LocQuality ($q$)       & 28.2 & 45.1 & 62.3 & 69.5 \\
        Aggregated ($o^{Agg}$) & 28.2 & 45.1 & 62.6 & 70.0 \\
        \bottomrule
    \end{tabular}
\end{table}

\begin{figure}[ht]
    \centering
    \includegraphics[width=0.9\linewidth]{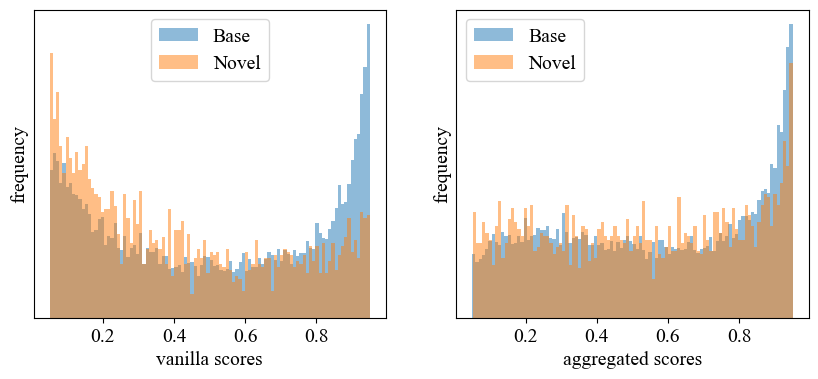}
    \caption{Statistical difference on predicted classification scores.}
    \label{fig:cls_stat}
\end{figure}

\textbf{Further ablations on object-classification aggregation. }
At the object-classification stage, we proposed to aggregate region-text similarities with region-prototype approximates, in order to adjust the underestimated classification scores on novel classes.
However, a naive alternative solution is to amplify the region-text similarity $s = \textbf{f} \cdot \textbf{t}$ by adding a trivial offset $\alpha_0$.
We particularly calculate the offset $\alpha_0$ by comparing the statistical difference between the vanilla $s$ and our $s^{Agg}$ on a set of randomly selected samples, and perform the naive aggregation to $s$.
However, as \cref{tab:res_ablations_cls} shows, such a trivial solution may improve the detection of novel classes, while significantly decreasing the performance of base classes.
The main reason is the amplification of all the novel categories introduces numerous false-positive predictions, leading to the inferiority of final performance.
On the contrary, our proposed aggregation $s^{Agg}$ effectively detects the novel classes (from 27.8\% to 28.6\%), while roughly keeping the performance on base classes (with a minor 0.1\% decline).
In addition, \cref{fig:cls_stat} displays the statistical difference between the classification scores on novel and base categories, and our method successfully aligns the underestimated confidence in novel classes with the well-estimated base scores.
It is one of the key points to improve the OVOD performance.

\begin{table}[ht]
\caption{Ablations on visual similarity aggregation for the object-classification stage.}
    \label{tab:res_ablations_cls}
    \centering
    \begin{tabular}{@{}c|ccc@{}}
        \toprule
        Confidence Estimate & $\text{mAP}_{\text{50}}^{\text{novel}}$ & $\text{mAP}_{\text{50}}^{\text{base}}$ & $\text{mAP}_{\text{50}}^{\text{all}}$ \\
        \midrule
        Baseline   ($s = \textbf{f} \cdot \textbf{t}$) & 27.8 & 51.1 & 45.0 \\
        Trivial    ($s + \alpha_0$)            & 28.8 & 49.9 & 44.4 \\
        Aggregated ($s + \alpha * \textbf{f} \cdot \textbf{p}$) & 28.6 & 51.0 & 45.1  \\
        \bottomrule
    \end{tabular}
\end{table}

\begin{figure}[ht]
    \centering
    \includegraphics[width=1.0\linewidth]{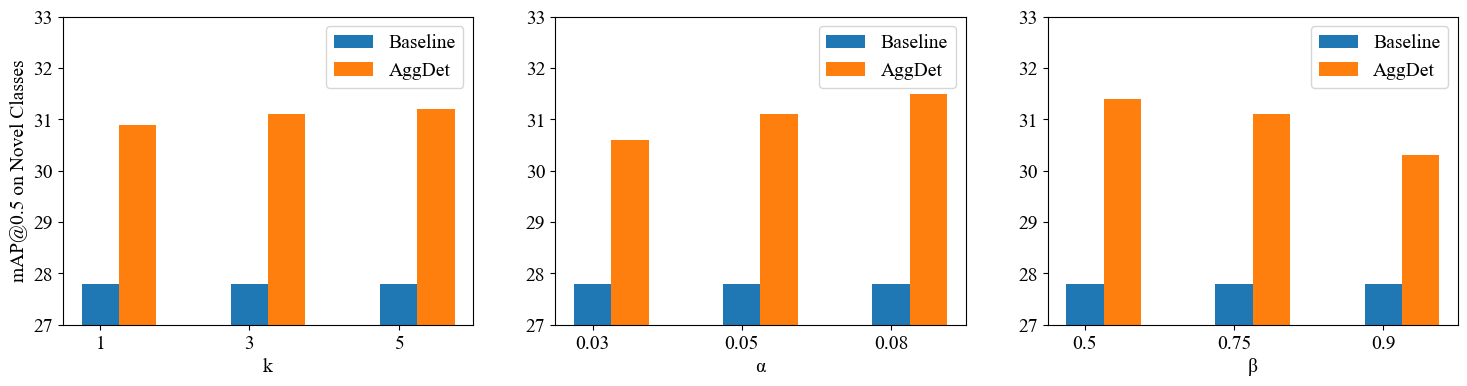}
    \caption{Ablation on hyper-parameter sensitivity.}
    \label{fig:sensitive}
\end{figure}

\textbf{Parameters stability. }
Furthermore, we verified the stability of our proposed method concerning hyper-parameter variations. 
In \cref{fig:sensitive}, we individually perturb the hyper-parameters in a pre-defined range regarding the TopK overlaps to estimate the localization quality ($k$), and the scaling factors for object-classification aggregation with visual similarities ($alpha$) and localization quality ($\gamma$). 
The remaining parameters were kept at their default values in \cref{tab:hyper}. 
According to the detection performance on novel classes in \cref{fig:sensitive}, the consistent improvements demonstrate the parameter stability of our \mmethod. 

\textbf{Time consumption. } 
On an RTX 3090 GPU, we report the extra time consumption introduced by our method.
As \cref{tab:time_cost} shows, \mmethod~introduced a negligible 0.7 ms cost during the inference, as we only employ the highly optimized addition, multiplication, and TopK selection operations.
The practicality of our method is resolved.

\begin{table}[ht]
\caption{Measurements on the newly introduced time consumption.}
    \label{tab:time_cost}
    \centering
    \begin{tabular}{@{}ccc@{}}
        \toprule
        Vanilla Cost & Introduced Cost & Gain Ratio \\
        \midrule
        70.1 ms & 0.7 ms & 1.0\% \\
        \bottomrule
    \end{tabular}
\end{table}
\section{Conclusion}
\label{sec:conclu}

In the open-vocabulary object detection setting, his paper systematically investigates the underestimated object for novel classes in commonly employed two-stage detectors.
Consequently, a generic and training-free post-processing schema is proposed to aggregate confidence, by using specifically designed techniques for both two detection stages.
The ultimate \mmethod~showcase its advances and generalization abilities with extensive experiments and ablations, which bolsters various OVOD detectors across model scales and architecture designs, and is able to further promote SOTA models.
We hope this paper can bring new insights to the open-vocabulary object detection community.

\clearpage  

%
%
\bibliographystyle{splncs04}
\bibliography{content/reference}


\newpage
\appendix

\renewcommand\thesection{\Alph{section}}
\renewcommand\thefigure{A\arabic{figure}}    
\renewcommand\thetable{A\arabic{table}} 
\setcounter{figure}{0}  
\setcounter{table}{0}  

\section{Discussion on the Visual Prototypes}
\label{sec:app_A}
At \cref{sec:method_obj} in our manuscript, we randomly select 300 samples from the training set to calculate the visual prototypes for each category. 
In this section, we conduct more experiments to investigate the way to compute these prototypes, which mainly comprise two aspects:

\begin{itemize}
    \item \textbf{Random Sampling}. For simplicity, we randomly select $N$ samples from the training set, extract their visual embeddings, and average them to derive the visual prototype for each class. We vary the values of $N$ to study the effect of sample size on prototype quality.

    \item \textbf{Top-$K$ Sampling}. Unlike random sampling, we leverage the confidence scores from the base OVOD detectors (\eg, Detic~\cite{zhou2022detic}) as the selection criteria. We first score all the training samples using the baseline detector, select the top-$K$ highest-scoring samples for each class, and average them for the visual prototypes. We experiment with different values of $K$ to evaluate its effect.
\end{itemize}

The results in \cref{tab:res_sample} illustrate that different computation strategies bring consistent improvement for open-vocabulary object detection, irrespective of the particularities of parameter configurations. 
Randomly selecting 300 samples and averaging them produces sufficiently robust visual prototypes.

\begin{table}[ht]
\caption{Performance comparison of computation strategies for visual prototypes on OV-COCO~\cite{lin2014microsoft}. The base detector is Detic\cite{zhou2022detic} with a ResNet-50 backbone.}
    \label{tab:res_sample}
    \centering
    \begin{tabular}{@{}c|ccc@{}}
        \toprule
        Sampling Strategy & $\text{mAP}_{\text{50}}^{\text{novel}}$ & $\text{mAP}_{\text{50}}^{\text{base}}$ & $\text{mAP}_{\text{50}}^{\text{all}}$  \\
        \midrule
        None       & 27.8 & 51.1 & 45.0 \\
        \midrule
        Random ($N=100$)   & 31.1 & 51.1 & 45.9 \\
        Random ($N=300$)   & 31.1 & 51.1 & 45.9  \\
        Random ($N=500$)   & 31.1 & 51.1 & 45.9 \\
        Top-$K$ ($K=100$)  & 31.0 & 51.2 & 45.9 \\
        Top-$K$ ($K=300$)  & 31.1 & 51.1 & 45.9 \\
        Top-$K$ ($K=500$)  & 31.1 & 51.1 & 45.9 \\
        \bottomrule
    \end{tabular}
\end{table}

\section{Visualization}
\label{sec:app_B}
\begin{figure}[ht]
    \centering
    \begin{subfigure}{1.0\linewidth}
        \centering
        \includegraphics[width=1.0\linewidth]{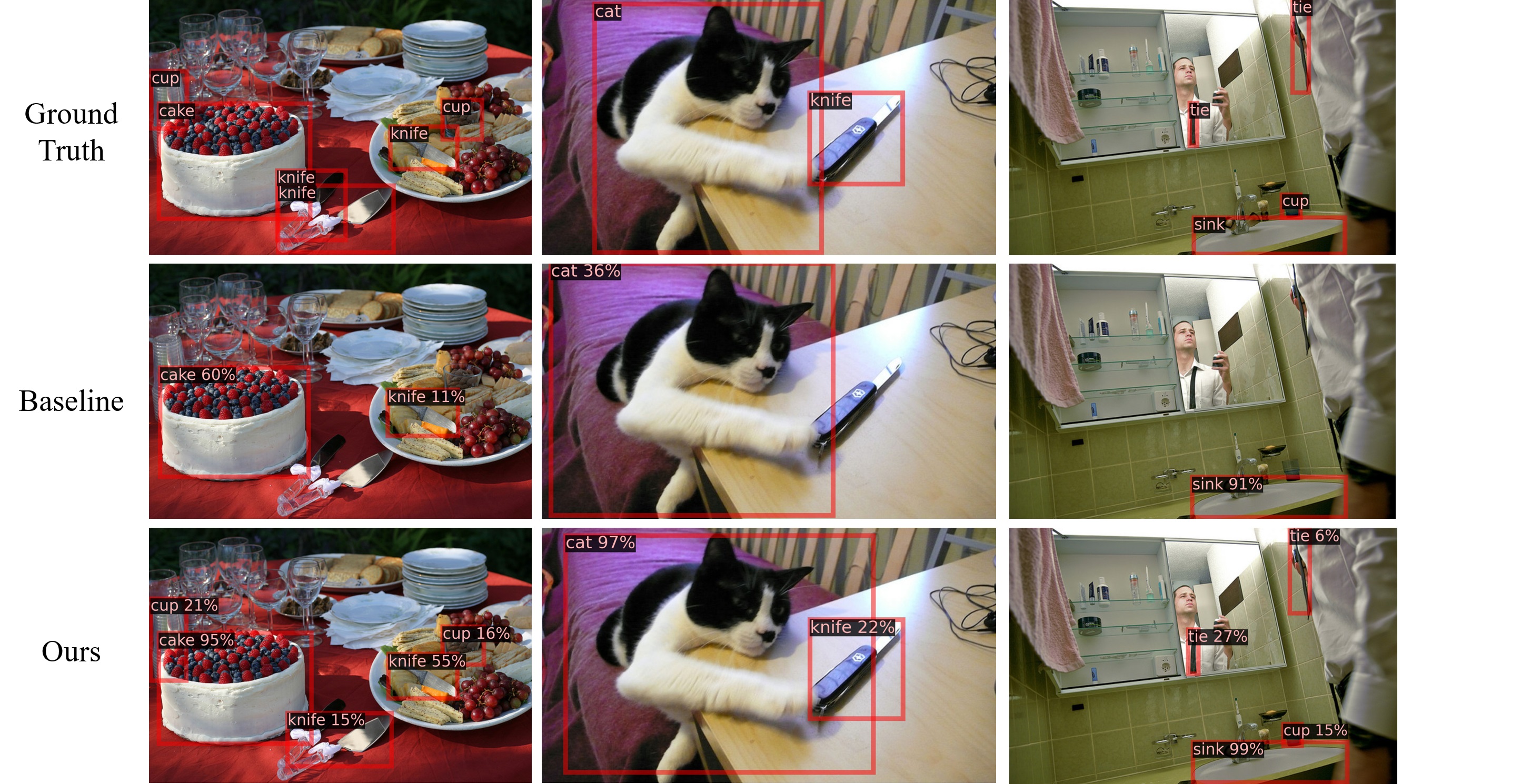}
        \caption{Detecting more novel instances that were missed by baseline detectors.}
    \end{subfigure}

    \vspace{10pt}

    \begin{subfigure} {1.0\linewidth}
        \centering
        \includegraphics[width=1.0\linewidth]{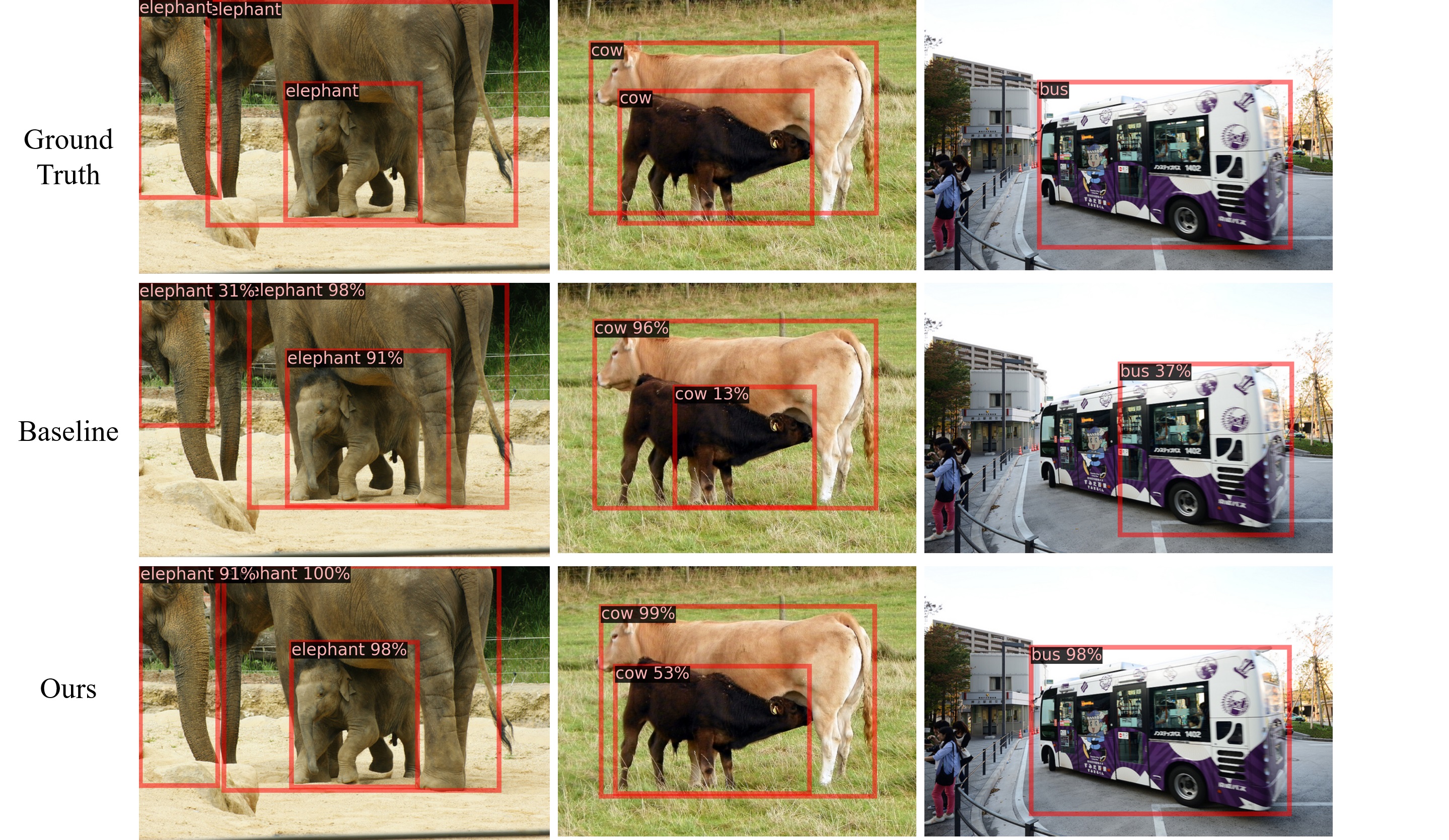}
        \caption{More precise bounding box predictions for novel instances.}
    \end{subfigure}
    
    \caption{Visualization examples on OV-COCO~\cite{lin2014microsoft} using Detic~\cite{zhou2022detic} with a ResNet-50 backbone. The first row shows the ground truth annotations, the second row displays the predictions of the baseline model, and the third row presents the enhanced predictions with our \mmethod.}
    \label{fig:a1}
\end{figure}

\begin{figure}[ht]
    \centering
    \begin{subfigure}{1.0\linewidth}
        \centering
        \includegraphics[width=1.0\linewidth]{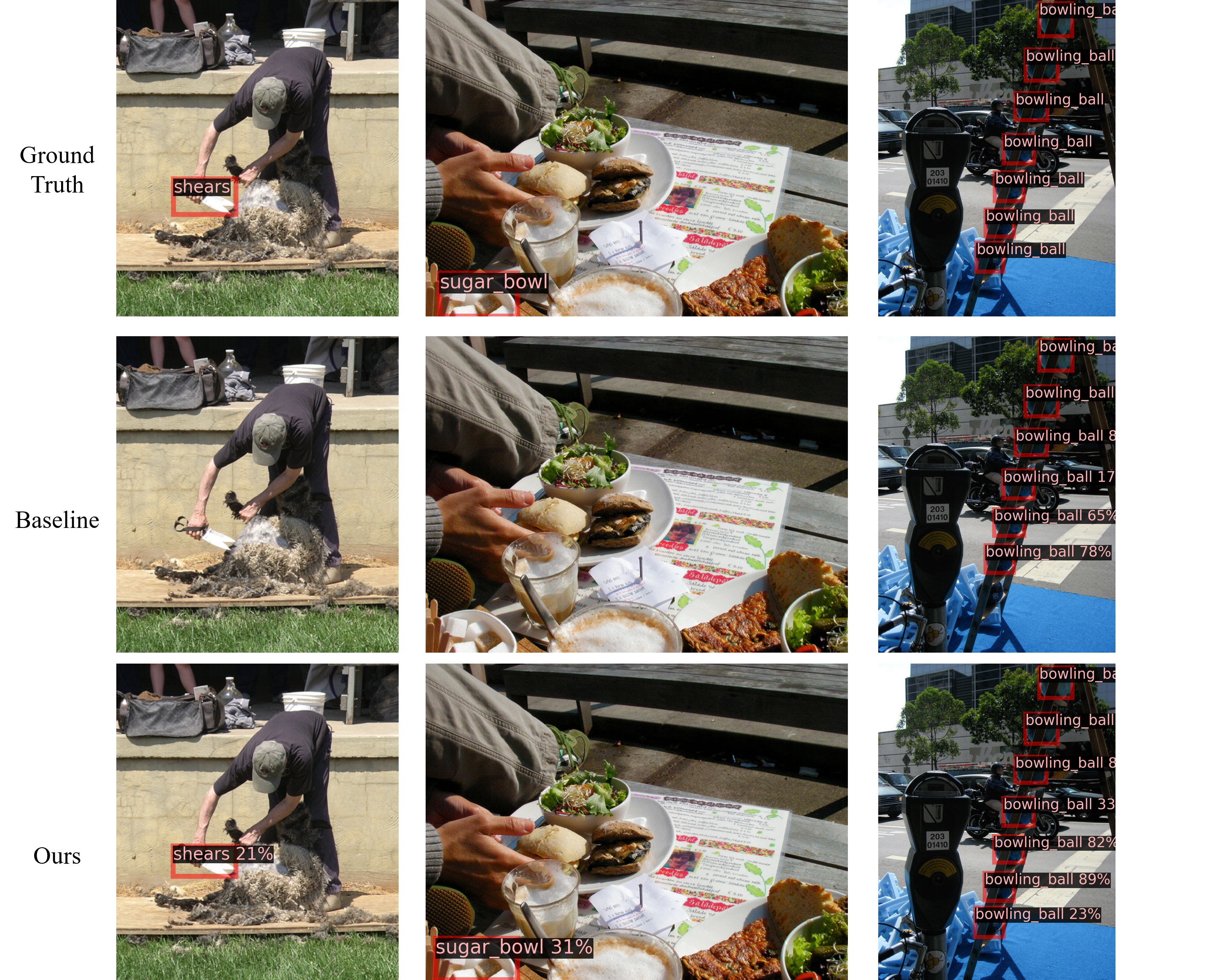}
        \caption{Boosting Rasheed\cite{bangalath2022bridging} with a ResNet-50 backbone.}
    \end{subfigure}

    \vspace{10pt}

    \begin{subfigure} {1.0\linewidth}
        \centering
        \includegraphics[width=1.0\linewidth]{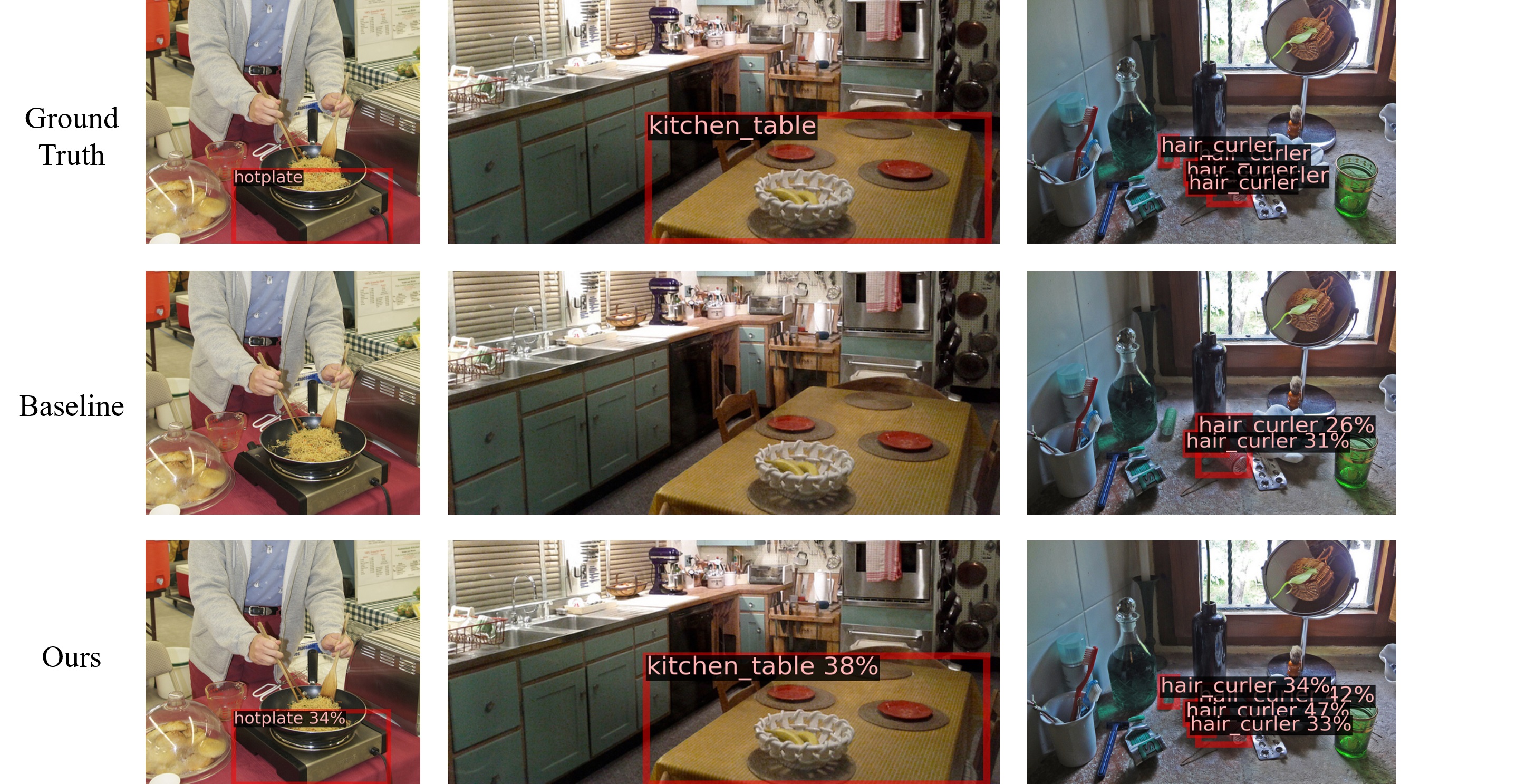}
        \caption{Boosting CoDet\cite{ma2024codet} with a Swin-B backbone.}
    \end{subfigure}
    
    \caption{Visualization examples on the OV-LVIS~\cite{gupta2019lvis} dataset. The first row shows the ground truth annotations, the second row displays the predictions of the baseline model, and the third row presents the enhanced predictions with our \mmethod.}
    \label{fig:a2}
\end{figure}

In this section, we provide some typical visualization examples, to delve into the efficacy of boosting open-vocabulary object detection by our \mmethod. 
As shown in \cref{fig:a1}, the performance gain of \mmethod~on novel classes primarily comes from two perspectives: (1) the ability to detect novel instances that the baseline model failed to recognize, and (2) more precise bounding box predictions for some novel instances. 
More visualization examples are shown in \cref{fig:a2}.

\end{document}